\pdfoutput=1
\documentclass[11pt]{article}

\usepackage[preprint]{acl}

\usepackage{times}
\usepackage{latexsym}

\usepackage[T1]{fontenc}

\usepackage[utf8]{inputenc}

\usepackage{microtype}

\usepackage{inconsolata}

\usepackage{graphicx}
\usepackage{multirow}
\usepackage{amsmath}
\usepackage{algorithm}
\usepackage{algorithmic}
\usepackage{float}
\usepackage{enumitem}
\usepackage[most]{tcolorbox}
\title{SAVER: Selective Auditing of Verbal Evidence for Error Recovery in VLM Change Reasoning}

\author{Youdi Li \\
  Panasonic Connect Co., Ltd. \\
  Tokyo, Japan \\
  \texttt{ri.yutei@jp.panasonic.com}}

\begin{document}
\maketitle
\begin{abstract}
Vision-language models (VLMs) frequently fail at visual change reasoning, even when their vision encoders contain sufficient information. We observe that correct VLM outputs tend to contain explicit verbal evidence (object names, colors, spatial locations) that supports the claimed change, while incorrect outputs often lack such evidence. We propose SAVER (Selective Auditing of Verbal Evidence for Error Recovery), a lightweight, rule-based method that parses VLM responses for this evidence and triggers structured reprompting only when evidence is missing or inconsistent. Across three change detection benchmarks and four VLMs, SAVER significantly improves accuracy on tasks where errors stem from the model failing to articulate what it saw (expression failures), with gains up to $+$25.8\% on CLEVR-Change. The evidence patterns can also be generated by an LLM in a single call, matching the hand-tuned gate on CLEVR-Change. Ablation experiments confirm that the evidence gate, not reprompting alone, drives the improvement.
\end{abstract}

\section{Introduction}
\label{sec:intro}

\begin{figure}[!ht]
  \centering
  \includegraphics[width=\columnwidth]{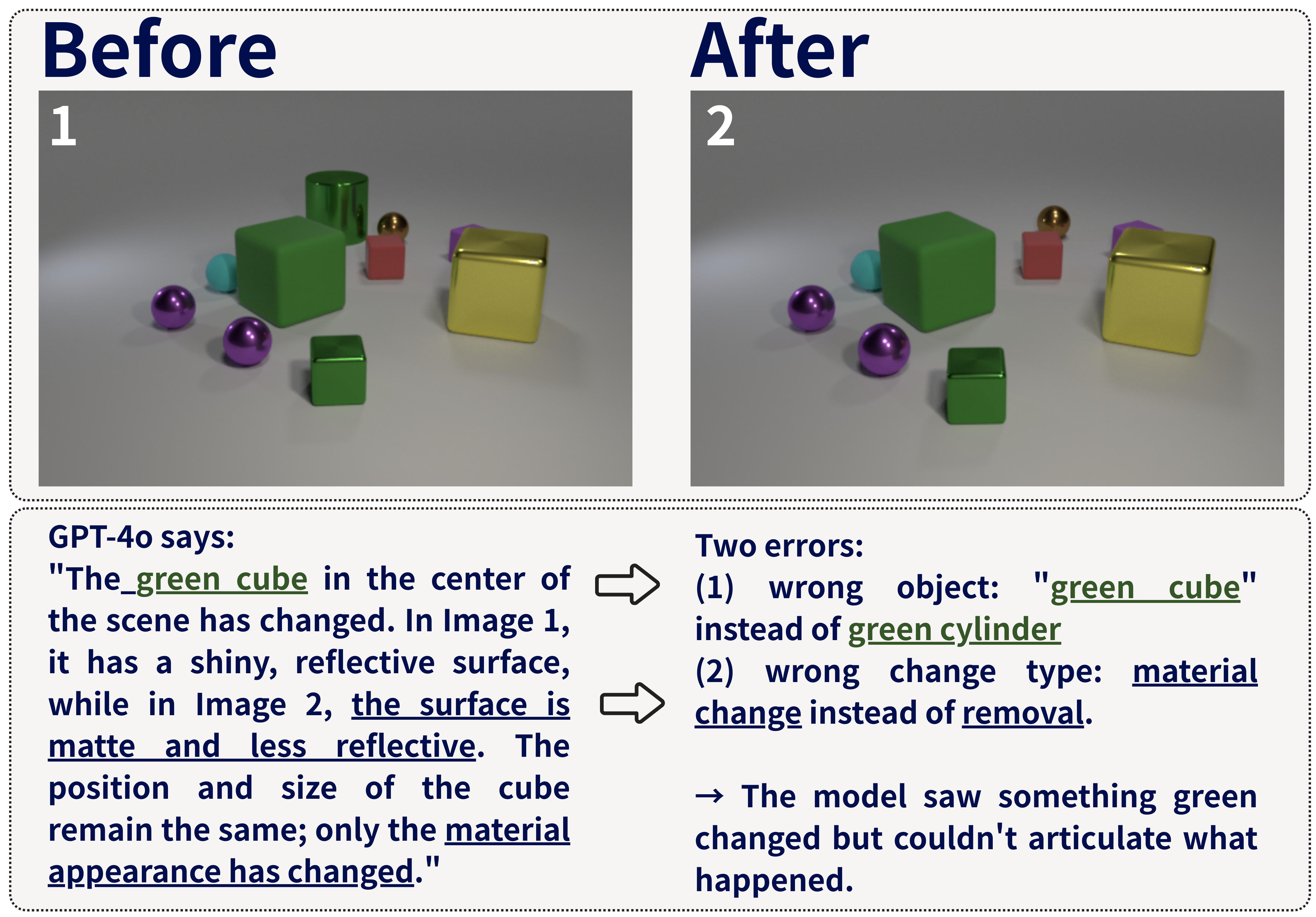}
  \caption{Expression failure in VLM change reasoning (one sample from CLEVR-Change). The ground truth is removal of the green cylinder, but GPT-4o misidentifies both the object and the change type. The model detects a change in the green region but cannot express it correctly.}
  \label{fig:problem}
\end{figure}

Vision-language models (VLMs) fail at change detection. Given two images of a scene, they miss obvious differences, hallucinate changes that do not exist, or describe changes in vague terms that lack specific detail \cite{liu2025omnidiff,rahmanzadehgervi2024vision}. Figure~\ref{fig:problem} shows a typical failure: GPT-4o identifies the wrong object (``green cube'' instead of the green cylinder) and the wrong change type (material change instead of removal), despite the change being visually obvious.

Why do VLMs fail? Recent work points to a gap between perception and expression. \citet{rahmanzadehgervi2024vision} showed that VLMs fail on simple visual tasks despite the visual information being present in the input. \citet{orgad2025llms} showed that LLM internal representations encode truthfulness information not reflected in outputs. \citet{balasubramanian-etal-2025-closer} found that VLMs sometimes reason correctly toward the ground truth but then change their answer, a pattern they call \emph{inconsistent reasoning}. These findings suggest a common failure mode: the model processes the visual input correctly but fails to express the relevant evidence in its output.

We propose SAVER (\textbf{S}elective \textbf{A}uditing of \textbf{V}erbal \textbf{E}vidence for Error \textbf{R}ecovery), a lightweight method that uses this perception-expression gap at test time. SAVER sends a standard prompt and applies a rule-based \emph{evidence gate} that parses the response for key verbal evidence. If the required evidence is present and consistent with the claimed change type, the gate accepts the response. If evidence is missing or conflicting, the gate triggers a structured reprompt that guides the model through systematic comparison.

The evidence check itself adds no cost; only the reprompt does. Because the gate is a short list of keywords and rules, every decision can be inspected to see which evidence was missing and why a reprompt was issued. Unlike learned verification methods where decisions cannot be inspected \cite{khan2024consistency,reverse2025}, SAVER's gate logic is fully transparent.

SAVER is based on two observations about expression failures. First, correct and incorrect VLM outputs differ in a measurable way: correct answers contain explicit verbal evidence (object names, colors, spatial locations) that supports the claimed change, while incorrect answers often lack such evidence. This gap can be measured and varies by model and dataset (Section~\ref{sec:discussion:boundary}). Second, when a model produces an evidence-poor response, a structured reprompt that forces explicit object enumeration often recovers the correct answer. The model saw the relevant objects but did not express the right evidence until asked to enumerate systematically (Section~\ref{sec:experiments:pertype}).

We evaluate SAVER on three change detection benchmarks (CLEVR-Change \cite{park2019robust}, MagicBrush \cite{zhang2024magicbrush}, and Spot-the-Diff \cite{jhamtani-berg-kirkpatrick-2018-learning}) across four VLMs (GPT-4o, Gemini 2.0 Flash, Qwen2.5-VL-7B, and LLaMA-4-Scout). SAVER improves GPT-4o accuracy by +13.4\% on CLEVR-Change and +5.3\% on MagicBrush over the baseline, with larger gains for weaker models (up to +25.8\% for Qwen), while breaking 3--9.5$\times$ fewer correct answers than always-structured prompting. However, SAVER shows no significant effect on Spot-the-Diff. This pattern reflects the method's scope: selective reprompting helps when errors are primarily expressive, but not when they are primarily perceptual.

This work makes three contributions. First, we show that VLMs sometimes see the change but omit key evidence from their text output. This expression failure is systematic and detectable from the output text alone, without access to model internals. Second, we propose SAVER, a rule-based method that uses verbal evidence as a diagnostic signal to selectively trigger structured reprompting at test time. Third, we evaluate SAVER across three datasets and four models, showing when selective auditing helps (expression-dominant errors), when it does not (perception-dominant errors), that it consistently fixes more samples than it degrades, and that its patterns can be LLM-generated.

\section{Related Work}
\label{sec:related}

\paragraph{Image difference captioning.}
Introduced by \citet{jhamtani-berg-kirkpatrick-2018-learning} and established with CLEVR-Change \cite{park2019robust}, image difference captioning has driven specialized architectures \cite{hu2024onediff,evennou2025blip2idc,li-etal-2025-change,huang-etal-2025-image} and multi-image benchmarks showing that general-purpose VLMs lag well behind humans on cross-image comparison tasks \cite{zhang-etal-2025-vlm2}. These methods improve the model itself through training, while our work operates at inference time on frozen models, checking whether the output contains evidence that supports its claim.

\paragraph{Perception--expression gap in VLMs.}
Multiple lines of evidence show that VLM errors often originate not in perception but in generation. Vision encoders contain sufficient information that the language model fails to decode \cite{rahmanzadehgervi2024vision,tong2024eyes}. Internal representations encode correct answers that do not reach the output \cite{orgad2025llms,simhi2024distinguishing,balasubramanian-etal-2025-closer}. Reasoning traces can be unfaithful to the model's own knowledge \cite{turpin2023unfaithful,lanham2023measuring}. Benchmarks confirm the scale of the problem: GPT-4V scores 31.4\% on HallusionBench question pairs \cite{guan2024hallusionbench}, and GPT-4o achieves only 5.9 BLEU-4 on image difference captioning \cite{liu2025omnidiff}. \citet{liu2025seeing} showed that VLMs perceive visual evidence in intermediate representations but fail to use it during generation. We use this gap as a test-time signal: if the output lacks the evidence its claim requires, a targeted reprompt can recover it.

\paragraph{Selective verification.}
Always-verify strategies waste computation on inputs the model already handles \cite{snell2025scaling}. Recent methods implement selective triggers: dynamic early exit \cite{deer2025}, adaptive reasoning escalation \cite{arm2025}, and RL-based reasoning selection \cite{ton2025}. MM-Verify \cite{sun-etal-2025-mm} applies output-level parsing to mathematical reasoning. Methods closer to our setting each require resources beyond the output text: REVERSE \cite{reverse2025} needs access to token-level logits during generation, VERA \cite{chen2026vera} needs attention maps, ReCoVERR \cite{srinivasan-etal-2024-selective} deploys two additional models, and \citet{khan2024consistency} require multiple API calls for paraphrase consistency. In contrast, our approach parses only the response already produced, adding zero cost unless the gate triggers a reprompt.

\section{Method: SAVER}
\label{sec:method}

\begin{figure*}[!ht]
  \centering
  \includegraphics[width=\textwidth]{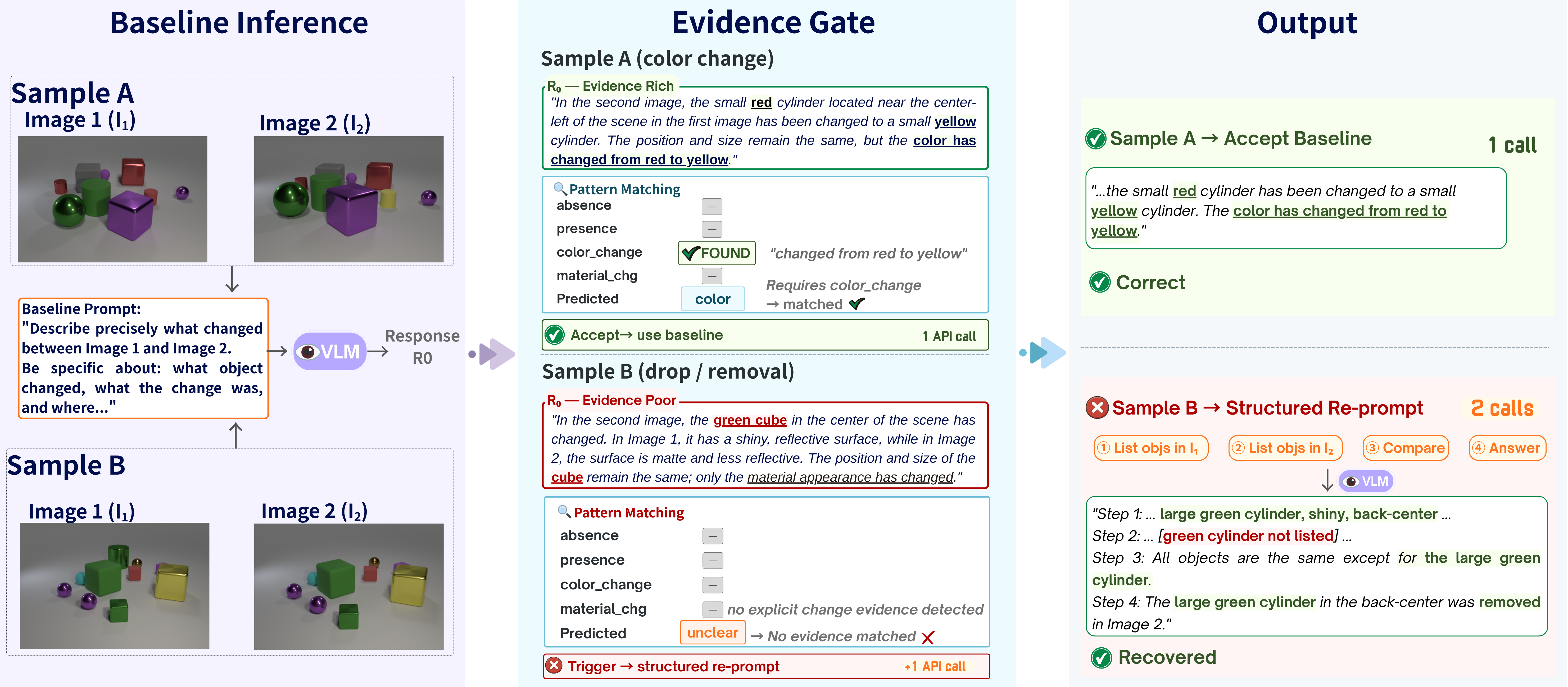}
  \caption{The SAVER pipeline. Given an image pair, SAVER sends a baseline prompt and parses the response for verbal evidence using rule-based pattern matching. \textbf{Sample~A} (color change): the response states ``changed from red to yellow,'' so the gate finds \texttt{color\_change} evidence matching the predicted type and accepts the baseline (1~call). \textbf{Sample~B} (drop): the baseline claims ``the green cube\ldots has changed'' and mentions material terms (``matte,'' ``reflective''), but no evidence vector registers a match; the response never makes an explicit change claim such as ``changed from matte to shiny.'' The predicted type is \texttt{unclear} and no evidence is matched, so the gate triggers. The structured reprompt forces object enumeration, and the model correctly identifies the removed cylinder (2~calls).}
  \label{fig:method}
\end{figure*}

SAVER is a test-time method that checks whether a VLM response contains the verbal evidence its own claim requires. Given a task with known evidence categories, the user defines keyword pattern sets and gate rules. SAVER sends a baseline prompt, parses the response for the specified evidence, and decides whether to accept or issue a structured reprompt. These evidence definitions can also be generated automatically by prompting an LLM with a task description and example outputs (evaluated in Section~\ref{sec:experiments:llmgen}). Once generated, the rules are applied deterministically at inference time with no LLM involvement in the gate itself. SAVER adds zero cost when evidence is already present. Figure~\ref{fig:method} illustrates the pipeline.

The underlying observation is that correct and incorrect VLM outputs differ systematically in verbal evidence. When a model answers correctly, it tends to state specific details that support its claim. When it errs, these details are often absent or inconsistent. We demonstrate SAVER on VLM change detection, where evidence categories (object names, colors, spatial terms) and gate rules are based on the structure of change types. The same approach extends to any domain with a structured taxonomy of expected outputs.

\subsection{Evidence Extraction}
\label{sec:method:extraction}

Evidence extraction converts the VLM's free-text response into a binary evidence vector using case-insensitive pattern matching. Let $\mathcal{P} = \{\mathcal{P}_1, \ldots, \mathcal{P}_K\}$ denote the keyword pattern sets for $K$ evidence categories. Given a response $r$, we extract a binary vector $\mathbf{e}(r) \in \{0,1\}^K$:
\begin{equation}
  e_k(r) = \begin{cases} 1 & \text{if } \exists\, p \in \mathcal{P}_k \text{ s.t.\ } p \text{ matches } r \\ 0 & \text{otherwise} \end{cases}
  \label{eq:evidence}
\end{equation}

Each category captures a type of visual evidence relevant to the dataset. For CLEVR-Change ($K{=}4$), the categories are: \texttt{absence} (e.g., \emph{missing}, \emph{removed}), \texttt{presence} (e.g., \emph{new}, \emph{added}), \texttt{color\_change} (change verbs combined with color terms), and \texttt{material\_change} (change verbs combined with material terms). MagicBrush uses five categories. It keeps \texttt{absence}, \texttt{presence}, and \texttt{color\_change}, drops \texttt{material\_change}, and adds \texttt{replacement} and \texttt{attribute\_change}. Spot-the-Diff contains multiple simultaneous changes per image pair with no fixed type taxonomy, so the gate checks only whether the response mentions any change evidence at all rather than matching evidence to a specific change type. The complete keyword lists are in Appendix~\ref{app:keywords}.

In addition to evidence vectors, the gate extracts a predicted change type $\hat{t}(r) \in \mathcal{T} \cup \{\texttt{unclear}\}$, where $\mathcal{T}$ is the set of valid change types for the dataset (e.g., $\mathcal{T} = \{\texttt{drop}, \texttt{add}, \texttt{color}, \texttt{texture}\}$ for CLEVR-Change). The type is determined by the same keyword matching used for evidence extraction: if the response contains predefined removal phrases (e.g., ``was removed,'' ``no longer present''), the predicted type is \texttt{drop}; if it contains addition phrases (e.g., ``was added,'' ``appeared''), the predicted type is \texttt{add}; and so on. If no type-indicating phrase is found, or if the response matches multiple conflicting types, the predicted type defaults to \texttt{unclear}.

\subsection{Gate Decision}
\label{sec:method:gate}

For each change type $t \in \mathcal{T}$, we define a required evidence index $\mathrm{req}(t) \in \{1, \ldots, K\}$ and a conflicting evidence index $\mathrm{conf}(t) \in \{1, \ldots, K\}$. Table~\ref{tab:gate_mapping} shows these mappings for CLEVR-Change.

\begin{table}[t]
  \centering
  \small
  \begin{tabular}{@{}lll@{}}
    \hline
    $\hat{t}$ & $\mathrm{req}(\hat{t})$ & $\mathrm{conf}(\hat{t})$ \\
    \hline
    \texttt{drop}    & \texttt{absence}         & \texttt{presence} \\
    \texttt{add}     & \texttt{presence}         & \texttt{absence} \\
    \texttt{color}   & \texttt{color\_change}    & \texttt{material\_change} \\
    \texttt{texture} & \texttt{material\_change} & \texttt{color\_change} \\
    \hline
  \end{tabular}
  \caption{CLEVR-Change gate mappings. Each predicted type requires specific evidence and conflicts with a different category.}
  \label{tab:gate_mapping}
\end{table}

The gate maps $(\hat{t}, \mathbf{e})$ to a binary decision:
\begin{equation}
  G(r) = \begin{cases}
    \textsc{trigger} & \text{if } \hat{t}(r) = \texttt{unclear} \\
    \textsc{trigger} & \text{if } e_{\mathrm{req}(\hat{t})} = 0 \\
    \textsc{trigger} & \text{if } e_{\mathrm{conf}(\hat{t})} = 1 \\
    \textsc{accept}  & \text{otherwise}
  \end{cases}
  \label{eq:gate}
\end{equation}

The gate triggers when (1) no change type can be identified, (2) the predicted type's required evidence is absent, or (3) conflicting evidence is detected. If none of these conditions holds, the baseline response is accepted. MagicBrush and Spot-the-Diff use analogous mappings adapted to their change types (Appendix~\ref{app:gate_rules}).

\subsection{Selective Reprompting}
\label{sec:method:reprompt}

When the gate triggers, SAVER sends a structured reprompt that guides the model through systematic comparison: first enumerate objects in Image~1, then enumerate objects in Image~2, then compare systematically, and finally state the answer. This design draws on chain-of-thought prompting \cite{wei2022chain,kojima2022large}, which improves reasoning by eliciting intermediate steps, and extends it to multimodal settings where decomposing visual input into explicit object descriptions before comparison has been shown to improve compositional reasoning \cite{zhang2024multimodal,mitra2024compositional}. Ablation C4 confirms that this structure is the active ingredient: a generic retry prompt with the same gate timing fails to replicate the improvement (Section~\ref{sec:experiments:ablation}). The structured prompt costs one additional API call. The full prompts are in Appendix~\ref{app:prompts}.

Given an image pair $x = (I_1, I_2)$, the final output is:
\begin{equation}
  \mathrm{SAVER}(x) = \begin{cases}
    r_0 & \text{if } G(r_0) = \textsc{accept} \\
    r_1 & \text{if } G(r_0) = \textsc{trigger}
  \end{cases}
  \label{eq:saver}
\end{equation}
where $r_0 = \mathrm{VLM}(x, p_{\text{base}})$ is the baseline response and $r_1 = \mathrm{VLM}(x, p_{\text{struct}})$ is the structured reprompt response. The expected API cost per sample is $1 + \tau$, where $\tau$ is the trigger rate. In most settings, the majority of samples require only the baseline call (Section~\ref{sec:experiments:results}). Algorithm~\ref{alg:saver} summarizes the full procedure.

\begin{algorithm}[t]
\caption{SAVER Inference Protocol}
\label{alg:saver}
\begin{algorithmic}[1]
\REQUIRE Image pair $x$, VLM $\mathcal{M}$, patterns $\mathcal{P}$, mappings $\mathrm{req}(\cdot)$, $\mathrm{conf}(\cdot)$
\ENSURE Final response $r^*$
\STATE $r_0 \leftarrow \mathcal{M}(x,\; p_{\text{base}})$ \hfill $\triangleright$ Baseline call
\STATE $\mathbf{e} \leftarrow \mathrm{Extract}(r_0,\; \mathcal{P})$ \hfill $\triangleright$ Eq.~\ref{eq:evidence}
\STATE $\hat{t} \leftarrow \mathrm{TypeExtract}(r_0)$
\IF{$\hat{t} = \texttt{unclear}$}
    \STATE $G \leftarrow \textsc{trigger}$
\ELSIF{$e_{\mathrm{req}(\hat{t})} = 0$}
    \STATE $G \leftarrow \textsc{trigger}$ \hfill $\triangleright$ Missing evidence
\ELSIF{$e_{\mathrm{conf}(\hat{t})} = 1$}
    \STATE $G \leftarrow \textsc{trigger}$ \hfill $\triangleright$ Conflicting evidence
\ELSE
    \STATE $G \leftarrow \textsc{accept}$
\ENDIF
\IF{$G = \textsc{trigger}$}
    \STATE $r^* \leftarrow \mathcal{M}(x,\; p_{\text{struct}})$ \hfill $\triangleright$ Structured reprompt
\ELSE
    \STATE $r^* \leftarrow r_0$ \hfill $\triangleright$ Keep baseline
\ENDIF
\RETURN $r^*$
\end{algorithmic}
\end{algorithm}

\subsection{Case Study}
\label{sec:method:example}

We trace SAVER through two CLEVR-Change samples to illustrate the gate logic.

\paragraph{Sample A (gate accepts).}
A color change: a small red cylinder turns yellow (upper path in Figure~\ref{fig:method}). GPT-4o responds: ``\emph{the small red cylinder has been changed to a small yellow cylinder. The color has changed from red to yellow.}'' The gate scores the predicted type as \texttt{color} and scans for \texttt{color\_change} evidence. The phrase ``from red to yellow'' is a specific color transition claim. Required evidence is present with no conflict, so the gate returns \textsc{accept}. Cost: 1 API call.

\paragraph{Sample B (gate triggers).}
A green cylinder is removed (lower path in Figure~\ref{fig:method}), but GPT-4o responds: ``\emph{the green cube in the center of the scene has changed\ldots the surface is matte and less reflective\ldots only the material appearance has changed.}'' The response uses material-related vocabulary (``matte,'' ``reflective'') but never makes an explicit change claim that would register as evidence (e.g., ``changed from matte to shiny''). No evidence vector is matched, so the predicted type is \texttt{unclear} and the gate returns \textsc{trigger}. The structured reprompt forces GPT-4o to enumerate all objects in both images. It responds: ``\emph{The large green cylinder in the back-center was removed in Image~2.}'' Cost: 2 API calls.

The general principle behind both cases is the same: the gate does not judge whether the answer is correct, but whether the response contains the type of evidence that a correct answer would require. A color change claim needs a color transition phrase; a removal claim needs absence language; a texture change claim needs material terms in a change context. For example, the \texttt{material\_change} pattern requires a change verb (e.g., ``changed,'' ``switched'') to appear near a material term (e.g., ``matte,'' ``shiny'') within a short text window. In Sample~B, the words ``matte'' and ``reflective'' appear, but never close to a change verb, so the pattern does not match and the evidence vector remains zero. Because no evidence category is matched, the predicted type defaults to \texttt{unclear} and the gate triggers. This design means the gate catches responses that \emph{sound} relevant but fail to commit to a specific, checkable claim. The reprompt then forces the model to enumerate objects and compare them, producing the structured evidence that the baseline lacked.

\section{Experiments}
\label{sec:experiments}

\subsection{Experimental Setup}
\label{sec:experiments:setup}

\paragraph{Datasets.}
We evaluate on three benchmark datasets spanning synthetic, edited, and natural images (Table~\ref{tab:datasets}): CLEVR-Change, MagicBrush, and Spot-the-Diff. CLEVR-Change and MagicBrush define fixed change-type taxonomies with four and five categories respectively. Our MagicBrush sample contains no replace instances (attribute 100, add 97, remove 94, color 49). In contrast, Spot-the-Diff contains multiple simultaneous changes without a predefined taxonomy and is evaluated on recall: how many of the true changes the model correctly mentions.

\paragraph{Models.}
We test four VLMs: GPT-4o \cite{openai2024gpt4o}, Gemini 2.0 Flash \cite{google2024gemini}, Qwen2.5-VL-7B \cite{qwen2025qwen25vl}, and LLaMA-4-Scout \cite{meta2025llama4}. The latter three are accessed via OpenRouter,\footnote{\url{https://openrouter.ai}} an API aggregator for hosted models. All use temperature 0. To check whether SAVER generalizes beyond these four models, we also run Claude 3.5 Sonnet \cite{anthropic2024claude35} on a 200-sample CLEVR-Change subset, where SAVER improves over both baselines with zero Broke cases (Appendix~\ref{app:claude}).

\begin{table}[t]
  \centering
  \small
  \begin{tabular}{@{}llc@{}}
    \hline
    \textbf{Dataset} & \textbf{Domain} & $N$ \\
    \hline
    CLEVR-Change  & Synthetic 3D    & 632 \\
    MagicBrush    & Edited photos   & 340 \\
    Spot-the-Diff & Natural photos  & 197 \\
    \hline
  \end{tabular}
  \caption{Dataset summary.}
  \label{tab:datasets}
\end{table}

\paragraph{Conditions.}
\begin{itemize}[nosep,leftmargin=*]
  \item \textbf{Baseline (B0):} direct prompt, one call.
  \item \textbf{Structured (B1):} enumerate-compare prompt, one call.
  \item \textbf{Standard CoT (B2):} ``think step by step,'' one call (CLEVR only).
  \item \textbf{SAVER:} B0 first; if the gate triggers, B1 follows (one or two calls).
\end{itemize}

\paragraph{Ablations.}
Four controls isolate the gate's and prompt's contributions:
\begin{itemize}[nosep,leftmargin=*]
  \item \textbf{Random (C1):} reprompts the same count as SAVER but at random.
  \item \textbf{Length (C2):} reprompts short responses.
  \item \textbf{Inverted (C3):} reprompts the opposite set from SAVER.
  \item \textbf{Generic reprompt (C4):} same gate as SAVER, but replaces structured enumeration with ``reconsider your previous answer.''
\end{itemize}
C1--C3 use cached responses (zero additional API cost). C4 requires new API calls (multi-turn: baseline then generic reprompt). We also compare against \textbf{Self-Consistency (SC-5)} \cite{wang2023selfconsistency}: sample the model 5 times at temperature 0.7, majority vote on detected change type, and return the longest response of the winning type.

\paragraph{Evaluation.}
Rule-based keyword matching: a response is correct if it names the correct change type and at least one ground-truth object keyword. We manually verified evaluation accuracy on 50 random CLEVR-Change samples and found 98\% agreement with human judgment. Note that gate keywords and evaluation keywords are different sets: the gate checks evidence \emph{categories} (e.g., color change verbs) while evaluation checks \emph{specific ground-truth terms} (e.g., exact colors). See Limitations for further discussion. We report bootstrap 95\% CIs and Bonferroni-corrected McNemar's tests.

\subsection{Main Results}
\label{sec:experiments:results}

\paragraph{CLEVR-Change.}
Table~\ref{tab:clevr_results} shows accuracy across all four models. SAVER improves over baseline (B0) for all four models (all $p < 0.001$). The effect is largest for Qwen, which has the lowest baseline accuracy and the highest trigger rate, because its baseline outputs frequently lack explicit evidence. For GPT-4o, SAVER outperforms B1 by 5.5\%, because structured prompting hurts drop and add types while the gate preserves baseline answers for these types (Table~\ref{tab:clevr_pertype}). For Gemini and Qwen, B1 yields higher accuracy than SAVER, but also overwrites more correct baselines (Table~\ref{tab:fixbroke}); we detail this below. Standard CoT (B2) falls between B0 and B1 for three models, suggesting that ``think step by step'' provides some but not all of the benefit of explicit object enumeration. LLaMA is the exception: B2 (90.5\%) exceeds B1 (86.2\%). As the only mixture-of-experts model tested, LLaMA may respond better to open-ended reasoning than to rigid enumeration instructions.

\begin{table}[t]
  \centering
  \small
  \begin{tabular}{@{}lccccc@{}}
    \hline
    \textbf{Model} & \textbf{B0} & \textbf{B1} & \textbf{B2} & \textbf{SAVER} & \textbf{Trig.} \\
    \hline
    GPT-4o     & 72.8 & 80.7 & 77.5 & \textbf{\underline{86.2}} & 39\% \\
    Gemini     & 81.5 & \textbf{95.6} & 87.3 & \underline{93.0} & 36\% \\
    Qwen       & 56.5 & \textbf{87.0} & 73.4 & \underline{82.3} & 49\% \\
    LLaMA      & 82.3 & 86.2 & \textbf{90.5} & \underline{88.9} & 37\% \\
    \hline
  \end{tabular}
  \caption{CLEVR-Change accuracy (\%, $N=632$). \textbf{Bold} = best; \underline{underline} = SAVER. Trig.\ = gate trigger rate. All SAVER vs.\ B0: $p < 0.001$.}
  \label{tab:clevr_results}
\end{table}

\paragraph{MagicBrush.}
On real photographs with instruction-guided edits (Table~\ref{tab:mb_results}), three of four models improve significantly over baseline ($p < 0.05$). Gemini shows a non-significant gain, likely because its baseline outputs are already evidence rich on this dataset (trigger rate 11\%). B1 yields higher accuracy than SAVER for all four models, because most MagicBrush errors are perception failures where models misidentify the edited region rather than omit evidence. The gate's low trigger rates (11--38\%) confirm this: models produce evidence-rich but factually wrong descriptions, so the gate accepts responses that are confident but incorrect. However, B1's accuracy advantage again comes with higher breakage (Table~\ref{tab:fixbroke}).

\begin{table}[t]
  \centering
  \small
  \begin{tabular}{@{}lcccc@{}}
    \hline
    \textbf{Model} & \textbf{B0} & \textbf{B1} & \textbf{SAVER} & \textbf{Trig.} \\
    \hline
    GPT-4o     & 57.6 & \textbf{70.3} & \underline{62.9} & 22\% \\
    Gemini     & 67.9 & \textbf{79.7} & \underline{68.8} & 11\% \\
    Qwen       & 38.5 & \textbf{68.8} & \underline{56.2} & 38\% \\
    LLaMA      & 58.5 & \textbf{74.4} & \underline{63.5} & 30\% \\
    \hline
  \end{tabular}
  \caption{MagicBrush accuracy (\%, $N=340$). \textbf{Bold} = best; \underline{underline} = SAVER.}
  \label{tab:mb_results}
\end{table}

\paragraph{Statistical summary.}
All CLEVR-Change improvements are significant ($p < 0.001$, McNemar's test with Bonferroni correction). On MagicBrush, three of four models show significant gains ($p < 0.05$). Gemini's confidence interval crosses zero. Full bootstrap CIs are in Table~\ref{tab:significance} (Appendix~\ref{app:statistics}).

\paragraph{Spot-the-Diff.}
SAVER shows no significant improvement on this dataset for any model (all $p > 0.05$; Table~\ref{tab:spotdiff_full}). Trigger rates are low for GPT-4o, Gemini, and Qwen (12--17\%), because baseline responses contain change-related keywords even when incorrect: models describe plausible but wrong changes in detail, producing evidence-rich outputs that the gate accepts. LLaMA is an outlier with 57\% trigger rate but only a 12\% fix rate among triggered samples, indicating that even when the gate detects sparse outputs, the underlying errors are perceptual rather than expressive. We report this null result as an important scope limitation: SAVER's evidence gate is effective only when errors stem from expression failures, not perception failures. Appendix~\ref{app:spotdiff_cases} shows this directly: 27--46\% of gate-accepted responses are wrong despite containing evidence language. We analyze the boundary further in Section~\ref{sec:discussion:boundary}.

\begin{table}[t]
  \centering
  \small
  \begin{tabular}{@{}lcccc@{}}
    \hline
    \textbf{Model} & \textbf{B0} & \textbf{SAVER} & \textbf{$\Delta$} & \textbf{Trig.} \\
    \hline
    GPT-4o  & 77.2 & 78.2 & +1.0 & 14\% \\
    Gemini  & 67.5 & 70.1 & +2.5 & 17\% \\
    Qwen    & 61.4 & 62.9 & +1.5 & 12\% \\
    LLaMA   & 73.1 & 78.2 & +5.1 & 57\% \\
    \hline
  \end{tabular}
  \caption{Spot-the-Diff results (\% recall, $N=197$). All $p > 0.05$ (Bonferroni). Fix/Broke counts: GPT-4o 2/0, Gemini 5/0, Qwen 3/0, LLaMA 14/4.}
  \label{tab:spotdiff_full}
\end{table}

\paragraph{Robustness: Fix/Broke analysis.}
A reprompting strategy may degrade already-correct answers. We count \emph{Fix} (baseline wrong, method correct) and \emph{Broke} (baseline correct, method wrong). Table~\ref{tab:fixbroke} compares SAVER and B1. SAVER fixes more than it breaks for every model on every dataset, with Fix:Broke ratios from 2.5:1 to 74:1. B1 breaks 3--9.5$\times$ more correct baselines than SAVER, because B1 applies enumeration to \emph{all} samples including those where baseline is already correct. For GPT-4o on CLEVR-Change, B1 breaks 76 samples (12\%) while SAVER breaks only 8 (1.3\%). On MagicBrush, B1 breaks 14--21 correct baselines per model while SAVER breaks only 2--4. Even where B1 yields higher aggregate accuracy (Gemini and Qwen on CLEVR, all models on MagicBrush), it does so by overwriting many correct baselines that SAVER would have preserved. When degrading a correct answer is costlier than missing an improvement, selective gating is preferable.

\begin{table}[t]
  \centering
  \small
  \begin{tabular}{@{}llrrr@{}}
    \hline
    & \textbf{Model} & \textbf{Fix} & \textbf{Broke} & \textbf{B1 Broke} \\
    \hline
    \multirow{4}{*}{CLEVR}
    & GPT-4o & 93  & 8  & 76 \\
    & Gemini & 74  & 1  & 9  \\
    & Qwen   & 173 & 10 & 30 \\
    & LLaMA  & 59  & 17 & 67 \\
    \hline
    \multirow{4}{*}{MB}
    & GPT-4o & 21  & 3  & 21 \\
    & Gemini & 5   & 2  & 14 \\
    & Qwen   & 64  & 4  & 15 \\
    & LLaMA  & 20  & 3  & 18 \\
    \hline
  \end{tabular}
  \caption{Fix/Broke counts for SAVER and always-structured (B1) across CLEVR-Change and MagicBrush (MB). Fix and Broke are for SAVER; B1~Broke shows how many correct baselines B1 overwrites.}
  \label{tab:fixbroke}
\end{table}

\subsection{Per-Change-Type Analysis}
\label{sec:experiments:pertype}

To understand why SAVER outperforms B1 for GPT-4o, we examine the per-type breakdown (Table~\ref{tab:clevr_pertype}). Structured prompting helps color (+17\%) and texture (+49\%) but hurts drop ($-$22\%) and add ($-$13\%). We hypothesize that structured enumeration forces the model to list all objects, which for drop/add changes introduces confusion: the model lists objects present in both images and struggles to identify which one was added or removed. SAVER selectively applies structured reasoning where it helps, preserving baseline performance on drop/add while capturing most of the gains on color/texture.

The per-type results also reveal why the gate works. For drop changes, correct baselines typically contain absence evidence (``the red cube is no longer present''), which the gate detects and preserves. For color/texture changes, correct answers require specific property transitions (``changed from red to blue''), which baselines often replace with vague descriptions (``something changed about the sphere''). A key finding: in 68\% of GPT-4o cases where structured prompting corrected a baseline error (86/126), the baseline already identified the correct object but mischaracterized the change type. Texture changes dominate this pattern (59\% of mischaracterizations), as expected for visually subtle material changes. Similar rates hold for Gemini (62\%) and LLaMA (65\%). Qwen's lower rate (22\%) reflects more diverse errors from the smaller 7B model.

\begin{table}[t]
  \centering
  \small
  \begin{tabular}{@{}lcccc@{}}
    \hline
    \textbf{Condition} & \textbf{Drop} & \textbf{Add} & \textbf{Color} & \textbf{Texture} \\
    \hline
    B0 (baseline)  & 87.3 & 76.9 & 75.9 & 51.0 \\
    B1 (structured) & 65.6$^\dagger$ & 64.4$^\dagger$ & \textbf{93.0} & \textbf{100.0} \\
    SC-5           & \textbf{93.0} & \textbf{82.5} & 80.4 & 53.5$^\dagger$ \\
    SAVER          & \underline{89.2} & 76.9 & \underline{84.8} & \underline{94.3} \\
    \hline
  \end{tabular}
  \caption{CLEVR-Change per-type accuracy (\%, GPT-4o). \textbf{Bold} = best; \underline{underline} = SAVER; $^\dagger$ = below baseline. SC-5 = self-consistency ($k$=5).}
  \label{tab:clevr_pertype}
\end{table}

\subsection{Ablation: Does the Gate Matter?}
\label{sec:experiments:ablation}

The ablations test two dimensions: gate selection quality and reprompt content quality. SAVER outperforms all four ablations on both datasets (Table~\ref{tab:ablation}).

For gate selection, Random (C1) and Inverted (C3) gates fall 11.5\% and 17.8\% below SAVER on CLEVR respectively, with C3 scoring below baseline (68.4\% vs.\ B0 72.8\%). For reprompt content, C4 uses SAVER's identical gate but replaces the structured enumeration with a generic retry. C4 is significantly worse than SAVER in 6 of 12 model--dataset combinations (Bonferroni-corrected $p < 0.05$) and sometimes falls below baseline (e.g., GPT-4o CLEVR: 70.3\% vs.\ B0 72.8\%), showing that the structured enumeration prompt, not the second chance itself, drives the improvement.

This pattern holds across all four models on CLEVR. On MagicBrush, the Inverted gate slightly outperforms SAVER for Gemini and LLaMA by 2\%, consistent with the lower gate precision on real photographs discussed above. Appendix~\ref{app:ablation} reports three further gate variants (C5--C7): non-critical keyword triggers, a strict gate, and a loose gate. SAVER outperforms all three on CLEVR; on MagicBrush, the strict C6 gate performs slightly better. The self-consistency comparison is also in Appendix~\ref{app:ablation}.

\begin{table}[t]
  \centering
  \small
  \begin{tabular}{@{}lccc@{}}
    \hline
    \textbf{Condition} & \textbf{CLEVR} & \textbf{MB} & \textbf{Calls} \\
    \hline
    B0 (baseline)          & 72.8 & 57.6 & 1.0 \\
    B1 (always-structured) & 80.7 & 70.3 & 1.0 \\
    \hline
    \textbf{SAVER}         & \textbf{86.2} & \textbf{62.9} & 1.4 \\
    SC-5                   & 77.4 & --- & 5.0 \\
    C1: Random             & 74.7 & 60.3 & 1.4 \\
    C2: Length             & 69.0 & 61.2 & 1.4 \\
    C3: Inverted           & 68.4$^\dagger$ & 59.7$^\dagger$ & 1.4 \\
    C4: Generic reprompt   & 70.3$^\dagger$ & 54.7$^\dagger$ & 1.4 \\
    \hline
  \end{tabular}
  \caption{Ablation results (\% accuracy, GPT-4o). Calls = API calls per sample. MB = MagicBrush. \textbf{Bold} = best; $^\dagger$ = below baseline. SC-5 = self-consistency with $k$=5. C4 uses SAVER's gate but replaces the structured prompt with a generic retry.}
  \label{tab:ablation}
\end{table}

\subsection{LLM-Generated Evidence Patterns}
\label{sec:experiments:llmgen}

The manual gate requires hand-tuned keyword patterns. We test whether an LLM can generate them instead. For each dataset, we prompt Claude Sonnet 4.5 once with the task description, the change-type taxonomy, and 16 example baseline responses (8 correct, 8 incorrect, sampled with seed 42). The output is a complete pattern set in the same format as the manual gate. We use it as-is: no iteration, no manual edits. The generator is not among the four evaluated VLMs. We call this condition SAVER (LLM-gen). Table~\ref{tab:llmgen_results} shows the results for all 12 model--dataset cells.

\begin{table}[t]
  \centering
  \small
  \begin{tabular}{@{}llcccc@{}}
    \hline
    & \textbf{Model} & \textbf{B0} & \textbf{SAVER} & \textbf{SAVER-L} & \textbf{Trig.} \\
    \hline
    \multirow{4}{*}{CLEVR}
    & GPT-4o & 72.8 & 86.2 & \textbf{87.0} & 36\% \\
    & Gemini & 81.5 & \textbf{93.0} & 92.4 & 23\% \\
    & Qwen   & 56.5 & 82.3 & \textbf{84.2} & 65\% \\
    & LLaMA  & 82.3 & \textbf{88.9} & 88.8 & 35\% \\
    \hline
    \multirow{4}{*}{MB}
    & GPT-4o & 57.6 & \textbf{62.9} & 58.5 & 9\%  \\
    & Gemini & 67.9 & \textbf{68.8} & 67.4 & 7\%  \\
    & Qwen   & 38.5 & \textbf{56.2} & 40.6 & 14\% \\
    & LLaMA  & 58.5 & \textbf{63.5} & 61.2 & 26\% \\
    \hline
    \multirow{4}{*}{STD}
    & GPT-4o & 77.2 & \textbf{78.2} & 77.2 & 14\% \\
    & Gemini & 67.5 & \textbf{70.1} & 69.0 & 3\%  \\
    & Qwen   & 61.4 & \textbf{62.9} & 62.9$^\ddagger$ & 16\% \\
    & LLaMA  & 73.1 & \textbf{78.2} & 76.1 & 27\% \\
    \hline
  \end{tabular}
  \caption{SAVER (LLM-gen) results (\% accuracy). SAVER-L: SAVER with LLM-generated patterns. SAVER = manual patterns. Trig.\ = SAVER-L trigger rate. \textbf{Bold} = best. On CLEVR-Change, all SAVER-L vs.\ B0 gains are significant ($p < 0.001$, Bonferroni) and no SAVER-L vs.\ SAVER difference is significant. On MagicBrush, SAVER-L is significantly below SAVER for Qwen ($p < 0.001$, Bonferroni). $^\ddagger$ = lower bound: Qwen2.5-VL-7B was retired before these experiments, so 23 of 31 gate-triggered samples could not be reprompted and keep their baseline response. MB = MagicBrush; STD = Spot-the-Diff.}
  \label{tab:llmgen_results}
\end{table}

The result splits by dataset. On CLEVR-Change, single-shot generated patterns fully replace the manual gate for all four models. No difference from the manual gate is significant, and the generated gate is nominally better for GPT-4o and Qwen. The example responses shown to the generator came from GPT-4o logs only, so the patterns also transfer across models. On MagicBrush, the generated patterns fail for all four models. The generated \texttt{attribute\_change} patterns include bare terms like ``changed'' and ``different'' that match nearly any response, so missing evidence becomes undetectable and the gate under-triggers (trigger rates drop to 7--26\%, against 11--38\% for the manual gate). SAVER (LLM-gen) recovers only a fraction of the manual gain and is significantly below the manual gate for Qwen. On Spot-the-Diff, the generated gate changes nothing. The null result holds under both manual and generated patterns.

\paragraph{Self-critique pass.}
The MagicBrush failure traces to specific, identifiable patterns, the bare ``changed'' and ``different'' terms above. This raises a natural question: can the generator find and fix these patterns itself, without being told what went wrong? We test this with one additional step. The same LLM receives its own generated patterns and one generic instruction: flag any pattern broad enough to match almost any response, and tighten or remove it. The critique is a single pass. The LLM receives no accuracy feedback and no dataset-specific hints. Under this instruction, it removed exactly the bare ``changed'' and ``different'' patterns on its own. Table~\ref{tab:llmgen_critique} shows the effect.

\begin{table}[t]
  \centering
  \small
  \begin{tabular}{@{}lcccc@{}}
    \hline
    \textbf{Model} & \textbf{B0} & \textbf{SAVER} & \textbf{SAVER-L} & \textbf{+Critique} \\
    \hline
    GPT-4o & 57.6 & \textbf{62.9} & 58.5 & 60.0 \\
    Gemini & 67.9 & \textbf{68.8} & 67.4 & \textbf{68.8} \\
    Qwen   & 38.5 & \textbf{56.2} & 40.6 & 39.1 \\
    LLaMA  & 58.5 & \textbf{63.5} & 61.2 & \textbf{63.5} \\
    \hline
  \end{tabular}
  \caption{Self-critique on MagicBrush (\% accuracy). SAVER-L: SAVER with LLM-generated patterns. +Critique = the SAVER-L patterns after one self-critique pass. \textbf{Bold} = best. The critique restores manual-gate accuracy for Gemini and LLaMA, roughly halves the gap for GPT-4o, and fails for Qwen.}
  \label{tab:llmgen_critique}
\end{table}

The result is mixed. For Gemini and LLaMA, one critique pass restores manual-gate accuracy exactly. For GPT-4o, it roughly halves the gap (58.5 to 60.0, against 62.9 manual) and makes the gain over baseline significant ($p = 0.027$, uncorrected). For Qwen, it fails: the critiqued gate still triggers on only 10\% of samples, against 38\% for the manual gate, and accuracy stays near baseline. Qwen errs on 61\% of MagicBrush samples, and removing over-broad patterns cannot by itself restore a trigger rate that high. As a sanity check, we also ran the critique on the CLEVR-Change and Spot-the-Diff pattern sets. Accuracy is unchanged in seven of the eight cells and moves by $+$1.0 in one (Gemini on Spot-the-Diff, still not significant). The critique repairs broken patterns and does not harm working ones. We report single-shot and critiqued results as separate conditions throughout.

\section{Discussion}
\label{sec:discussion}

\paragraph{Expression vs.\ Perception Failures.}
\label{sec:discussion:taxonomy}
\label{sec:discussion:boundary}
SAVER's results across three datasets reveal two failure modes. \textbf{Expression failures} occur when the model captures the visual signal but omits key evidence from its output; the gate detects these and reprompting recovers the answer. \textbf{Perception failures} occur when the model misses the signal entirely, producing confident, evidence-rich but incorrect outputs that pass the gate. On CLEVR-Change, expression failures dominate (high trigger rates, high fix rates). On Spot-the-Diff, perception failures dominate (low trigger rates or low fix rates). This distinction determines when SAVER helps: the gate adds value when structured prompting helps some input types but hurts others (e.g., GPT-4o on CLEVR-Change), because it selectively applies structured reasoning only where evidence is missing. When errors are predominantly perceptual, B1 outperforms SAVER in aggregate accuracy because even evidence-rich responses may be wrong and the gate's selectivity provides no advantage. However, as the Fix/Broke analysis shows (Section~\ref{sec:experiments:results}), B1's accuracy comes at the cost of overwriting correct baselines at 3--9.5$\times$ the rate of SAVER. Which method to prefer depends on the deployment setting: B1 when aggregate accuracy is the only metric, SAVER when overwriting a correct answer carries a higher cost than missing an improvement. More generally, SAVER's type-specific gating is most effective when the task has a structured taxonomy of expected outputs and known evidence requirements (Appendix~\ref{app:domains}).

\paragraph{Measuring the boundary.}
This boundary can be measured. For each cell we compute the gate's precision at catching baseline errors and compare it to a random gate triggering at the same rate, whose precision equals the baseline error rate (Appendix~\ref{app:gate_precision}). The comparison separates the three datasets cleanly. The gate is above random on CLEVR-Change for all four models, at or below random on Spot-the-Diff, and in between on MagicBrush. The margin between a gate's own precision and the random baseline also predicts the end-to-end gain. For the 12 SAVER (LLM-gen) gates, the Spearman correlation between this margin (Table~\ref{tab:llmgen_precision}) and the gain over baseline is 0.68, rising to 0.74 when the self-critiqued gates are pooled in. A practitioner can therefore measure gate precision on a small labeled sample and predict whether SAVER will help before full deployment. The keyword quality analysis (Appendix~\ref{app:keyword_quality}) adds a second diagnostic. Pattern quality failures come in two kinds, and only one of them is dangerous. With conflict checks intact, over-broad patterns over-trigger and accuracy degrades gracefully toward always-structured prompting (B1). Without conflict checks, the trigger rate collapses silently and accuracy falls back to baseline, the same structure as the MagicBrush failure.

\paragraph{Failure modes.}
SAVER fails in two ways. First, when the model produces a confident but factually wrong response with correct-sounding evidence (perception failure), the gate accepts it. Second, on rare occasions the structured reprompt introduces errors: the forced enumeration step causes the model to second-guess correct baseline answers, particularly for drop/add types where listing all objects creates confusion about which one changed.

\section{Conclusion}
\label{sec:conclusion}

We introduced SAVER, a lightweight, rule-based method that checks VLM outputs for task-critical verbal evidence before deciding whether to trigger structured reprompting. SAVER significantly improves accuracy on expression-dominated tasks (up to $+$25.8\% on CLEVR-Change) while showing when and why the approach does not help (perception-dominated errors on Spot-the-Diff). Ablation experiments confirm that the evidence gate's selectivity, not reprompting alone, drives the improvement, and Fix/Broke analysis shows that SAVER degrades far fewer correct answers than always-structured prompting.

Our results reveal a practical lesson: VLMs often know more than they say, and the gap between perception and expression is structured enough to detect with simple keyword rules. Adapting SAVER to a new domain requires only specifying evidence patterns, which can be hand-written or LLM-generated and then checked (Section~\ref{sec:experiments:llmgen}). We anticipate the approach to be useful in high-stakes visual comparison settings where transparency and controllable verification are critical.

\section*{Limitations}
\label{sec:limitations}

The SAVER gate relies on predefined keyword vocabularies and a fixed proximity window for pattern matching that were set once and not tuned per dataset. Applying SAVER to a new domain requires defining new evidence patterns, and the patterns that work best likely differ across datasets and languages. However, the ablation results (Section~\ref{sec:experiments:ablation}) show that the gate's \emph{logic} matters more than any single keyword: random and inverted gates using the same keywords perform far worse. A second concern is circularity: the gate uses keyword matching and so does evaluation. We note that the gate checks \emph{evidence categories} (e.g., color change verbs like ``changed,'' ``switched'') while evaluation checks \emph{specific ground-truth terms} (e.g., exact colors like ``red,'' ``blue''). The two keyword sets share zero exact matches for CLEVR-Change: the gate's 23 patterns (Appendix~\ref{app:keywords}) contain no object names, colors, or shapes, while the evaluation vocabulary consists entirely of dataset-specific ground-truth terms. The inverted gate (C3) directly tests this: it uses the same 23 keywords but selects the opposite samples. If keyword overlap with evaluation drove the results, C3 should perform comparably to SAVER; instead it scores 17.8\% lower (68.4\% vs.\ 86.2\%), confirming that gate logic rather than keyword presence drives the improvement. Finally, although we discuss potential applications in high-stakes domains such as medical imaging (Appendix~\ref{app:domains}), SAVER should not be deployed in safety-critical settings without domain-specific validation. More fundamentally, SAVER addresses expression failures only. The gate accepts any response that contains the expected evidence patterns, so confident but factually wrong outputs pass unchecked when errors are perceptual, as the Spot-the-Diff null result illustrates. This scope boundary is measurable before deployment. On a small labeled validation set, gate precision must exceed the random-gate baseline (Appendix~\ref{app:gate_precision}), otherwise errors are not expressive and SAVER should not be expected to help.

Two further limitations come from the LLM-generated pattern experiments. First, one self-critique pass is not a universal fix. It restores manual-gate accuracy on MagicBrush for Gemini and LLaMA and halves the gap for GPT-4o, but it fails for Qwen2.5-VL-7B, where the critiqued gate still under-triggers. Second, the Spot-the-Diff SAVER (LLM-gen) result for Qwen2.5-VL-7B is a lower bound. The model was retired before the revision experiments, so 23 of its 31 gate-triggered samples could not be reprompted and keep their baseline responses.


\bibliography{references,custom}

\appendix

\section{Gate Decision Rules}
\label{app:gate_rules}

The gate maps extracted evidence vectors to a binary decision per dataset.

\paragraph{CLEVR-Change.}
The gate identifies the model's predicted change type, then checks:
\begin{itemize}
    \item \textbf{Drop} requires \texttt{absence} evidence. \texttt{presence} detected = conflict.
    \item \textbf{Add} requires \texttt{presence} evidence. \texttt{absence} detected = conflict.
    \item \textbf{Color} requires \texttt{color\_change} evidence. \texttt{material\_change} = conflict.
    \item \textbf{Texture} requires \texttt{material\_change} evidence. \texttt{color\_change} = conflict.
\end{itemize}

\paragraph{MagicBrush.}
Five change types map to required vectors: remove$\rightarrow$\texttt{absence}, add$\rightarrow$\texttt{presence}, color$\rightarrow$\texttt{color\_change}, replace$\rightarrow$\texttt{replacement}, attribute$\rightarrow$\texttt{attribute\_change}. Only remove and add have conflicting-evidence entries (\texttt{presence} and \texttt{absence} respectively). Color, replace, and attribute have none.

\paragraph{Spot-the-Diff.}
The gate checks whether the response contains \emph{any} change evidence. If the model reports ``no difference'' while mentioning specific changes, this conflict triggers a reprompt.

\section{Prompts}
\label{app:prompts}

We provide the exact prompts used for each dataset and condition. All prompts are sent with the image pair as visual input.

\subsection{CLEVR-Change Prompts}

\paragraph{Baseline (B0).}
\begin{quote}
\small
You are looking at two images of a scene with 3D objects. The second image shows the same scene after ONE change was made. Describe precisely what changed between Image 1 and Image 2. Be specific about: what object changed, what the change was (added, removed, moved, color changed, material changed), and where in the scene the change occurred.
\end{quote}

\paragraph{Structured (B1 / SAVER reprompt).}
\begin{quote}
\small
You are looking at two images of a scene with 3D objects. The second image shows the same scene after ONE change was made. Follow these steps carefully:

Step 1: Describe ALL objects in Image 1, listing each object's shape, color, material (shiny/matte), size (small/large), and approximate position.

Step 2: Describe ALL objects in Image 2, listing the same properties.

Step 3: Compare the two descriptions systematically to identify what changed.

Step 4: State your final answer: what object changed, what the change was (added, removed, moved, color changed, material changed), and where in the scene the change occurred.
\end{quote}

\paragraph{Standard CoT (B2).}
\begin{quote}
\small
You are looking at two images of a scene with 3D objects. The second image shows the same scene after ONE change was made. Think step by step about what changed between Image 1 and Image 2. Consider each object carefully and compare their properties. Then provide your final answer: what object changed, what the change was (added, removed, moved, color changed, material changed), and where in the scene the change occurred.
\end{quote}

\subsection{MagicBrush Prompts}

\paragraph{Baseline (B0).}
\begin{quote}
\small
You are looking at two images. The second image shows the same scene after ONE change was made. Describe precisely what changed between Image 1 and Image 2. Be specific about: what object changed, what the change was (added, removed, replaced, color changed, modified), and where in the scene the change occurred.
\end{quote}

\paragraph{Structured (B1 / SAVER reprompt).}
\begin{quote}
\small
You are looking at two images. The second image shows the same scene after ONE change was made. Follow these steps carefully:

Step 1: Describe the key objects and elements in Image 1, noting their appearance, color, position, and state.

Step 2: Describe the key objects and elements in Image 2, noting the same properties.

Step 3: Compare the two descriptions systematically to identify what changed.

Step 4: State your final answer: what object changed, what the change was (added, removed, replaced, color changed, modified), and where in the scene the change occurred.
\end{quote}

\subsection{Spot-the-Diff Prompts}

\paragraph{Baseline (B0).}
\begin{quote}
\small
You are looking at two images of the same scene taken at different times. Identify ALL differences between Image 1 and Image 2. For each difference, describe what changed: what object appeared, disappeared, moved, or changed in some way. Be specific about the objects and their locations.
\end{quote}

\paragraph{Structured (SAVER reprompt).}
\begin{quote}
\small
You are looking at two images of the same scene taken at different times. Follow these steps carefully:

Step 1: Describe the main objects and people visible in Image 1, noting their positions (left/right/center, foreground/background).

Step 2: Describe the main objects and people visible in Image 2, noting the same properties.

Step 3: Compare systematically --- for each object in Image 1, check if it exists in Image 2 and vice versa. Note any objects that appeared, disappeared, or moved.

Step 4: State your final answer: list ALL differences found, specifying for each: what object, what type of change (appeared, disappeared, moved, changed), and where in the scene.
\end{quote}

\subsection{Generic Reprompt (C4)}

The C4 ablation uses a multi-turn format: the baseline prompt and response are included as context, followed by this generic reprompt.

\begin{quote}
\small
I need you to reconsider your previous answer. Please look at the two images again carefully and provide a more thorough response.
\end{quote}

\section{Evidence Gate Keywords}
\label{app:keywords}

Table~\ref{tab:clevr_keywords} and Table~\ref{tab:mb_keywords} list the keyword patterns used by the SAVER evidence gate. All matching is case insensitive. Proximity patterns (marked $\sim$) match a change verb near a property term, allowing up to 20 intervening characters (e.g., ``changed the color'' matches \texttt{color\_change}).

\begin{table}[H]
  \centering
  \small
  \begin{tabular}{@{}lp{5cm}@{}}
    \hline
    \textbf{Vector} & \textbf{Keywords / Patterns} \\
    \hline
    \texttt{absence} & missing, absent, gone, disappeared, removed, no longer, not present \\
    \texttt{presence} & new, added, appeared, introduced, additional, placed \\
    \texttt{color\_change} & \{changed, switched, turned\} $\sim$ \{color\}; also detects color-pair mentions (e.g., ``red to blue'') \\
    \texttt{material\_change} & \{changed, switched\} $\sim$ \{material, texture, matte, shiny\} \\
    \hline
  \end{tabular}
  \caption{CLEVR-Change evidence keywords.}
  \label{tab:clevr_keywords}
\end{table}

\begin{table}[H]
  \centering
  \small
  \begin{tabular}{@{}lp{4.7cm}@{}}
    \hline
    \textbf{Vector} & \textbf{Keywords / Patterns} \\
    \hline
    \texttt{absence} & missing, absent, gone, disappeared, removed, no longer, not present, deleted, taken away, erased \\
    \texttt{presence} & new, added, appeared, introduced, additional, placed, now has/contains/includes/shows \\
    \texttt{color\_change} & \{changed, different, switched, turned, became\} $\sim$ \{color, colour, hue\}; also from-to patterns \\
    \texttt{replacement} & replaced, substituted, swapped, changed \ldots\ to/into/for/with \\
    \texttt{attribute\_change} & modified, altered, transformed, different \ldots\ shape/size/position \\
    \hline
  \end{tabular}
  \caption{MagicBrush evidence keywords.}
  \label{tab:mb_keywords}
\end{table}

\section{Statistical Details}
\label{app:statistics}

\begin{figure}[H]
  \centering
  \includegraphics[width=\columnwidth]{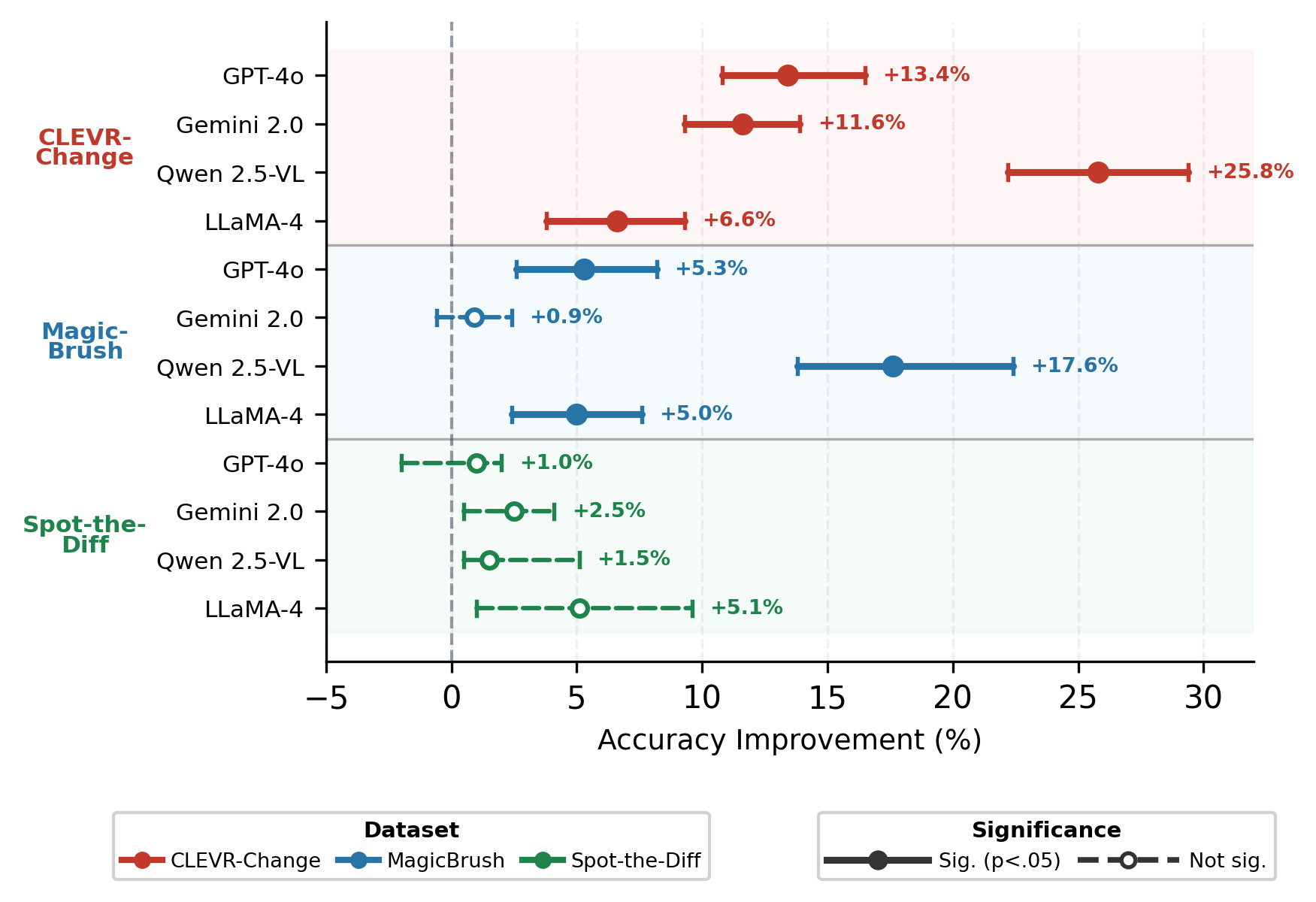}
  \caption{SAVER improvement over B0 (baseline) in percentage points, with 95\% bootstrap CIs across all 12 model--dataset combinations. Solid/filled = significant ($p < 0.05$); dashed/hollow = non-significant.}
  \label{fig:forest}
\end{figure}

Table~\ref{tab:significance} reports bootstrap confidence intervals and effect sizes for all model--dataset pairs. Figure~\ref{fig:forest} shows the forest plot.

\begin{table}[H]
  \centering
  \small
  \begin{tabular}{@{}llcc@{}}
    \hline
    & \textbf{Model} & \textbf{$\Delta$\%} & \textbf{95\% CI} \\
    \hline
    \multirow{4}{*}{CLEVR}
    & GPT-4o & +13.4$^{***}$ & [10.8, 16.5] \\
    & Gemini & +11.6$^{***}$ & [9.3, 13.9]  \\
    & Qwen   & +25.8$^{***}$ & [22.2, 29.4] \\
    & LLaMA  & +6.6$^{***}$  & [3.8, 9.3]   \\
    \hline
    \multirow{4}{*}{MB}
    & GPT-4o & +5.3$^{**}$   & [2.6, 8.2]   \\
    & Gemini & +0.9          & [$-$0.6, 2.4]  \\
    & Qwen   & +17.6$^{***}$ & [13.8, 22.4] \\
    & LLaMA  & +5.0$^{**}$   & [2.4, 7.6]   \\
    \hline
  \end{tabular}
  \caption{SAVER vs.\ baseline ($\Delta$\% with bootstrap 95\% CIs). $^{***}$\,$p<0.001$; $^{**}$\,$p<0.05$; no mark = n.s. MB = MagicBrush. Cohen's $h$: 0.19--0.57 (CLEVR), 0.02--0.36 (MB).}
  \label{tab:significance}
\end{table}

\section{Gate Precision Analysis}
\label{app:gate_precision}

We evaluate how well the manual gate separates wrong baseline responses from correct ones. For each model--dataset cell we run the gate on the cached baseline responses and label each sample against baseline correctness. A true positive is a triggered sample whose baseline response is wrong. A false positive is a triggered sample whose baseline response is correct. Gate precision is the fraction of triggered samples that are true positives. The reference point is a random gate triggering at the same rate. Its precision equals the baseline error rate, denoted $P_{\text{rand}}$. A gate is informative only if its precision exceeds $P_{\text{rand}}$. This analysis uses existing logs and requires no new API calls.

Table~\ref{tab:gate_precision} shows the result. On CLEVR-Change, precision exceeds $P_{\text{rand}}$ by 0.08 to 0.26 absolute for all four models. On Spot-the-Diff, precision is at or below $P_{\text{rand}}$ for all four models. MagicBrush lies in between, with margins from $-$0.08 to $+$0.18. The cells where the gate beats the random baseline are the cells where SAVER helps end-to-end (Section~\ref{sec:discussion}).

\begin{table}[H]
  \centering
  \small
  \begin{tabular}{@{}llcccc@{}}
    \hline
    & \textbf{Model} & \textbf{Trig.} & \textbf{Prec.} & \textbf{Rec.} & $P_{\text{rand}}$ \\
    \hline
    \multirow{4}{*}{CLEVR}
    & GPT-4o & 0.39 & \textbf{0.508} & 0.733 & 0.272 \\
    & Gemini & 0.36 & \textbf{0.371} & 0.726 & 0.185 \\
    & Qwen   & 0.49 & \textbf{0.691} & 0.782 & 0.435 \\
    & LLaMA  & 0.37 & \textbf{0.255} & 0.536 & 0.177 \\
    \hline
    \multirow{4}{*}{MB}
    & GPT-4o & 0.22 & \textbf{0.500} & 0.257 & 0.424 \\
    & Gemini & 0.11 & 0.243 & 0.083 & 0.321 \\
    & Qwen   & 0.38 & \textbf{0.792} & 0.493 & 0.615 \\
    & LLaMA  & 0.30 & \textbf{0.446} & 0.319 & 0.415 \\
    \hline
    \multirow{4}{*}{STD}
    & GPT-4o & 0.14 & 0.143 & 0.089 & 0.228 \\
    & Gemini & 0.17 & 0.294 & 0.156 & 0.325 \\
    & Qwen   & 0.12 & 0.208 & 0.066 & 0.386 \\
    & LLaMA  & 0.57 & 0.239 & 0.509 & 0.269 \\
    \hline
  \end{tabular}
  \caption{Manual-gate precision and recall against the baseline-correctness oracle, all 12 cells. Trig.\ = trigger rate. $P_{\text{rand}}$ = precision of a random gate at the same trigger rate (the baseline error rate). \textbf{Bold} = precision above $P_{\text{rand}}$. MB = MagicBrush; STD = Spot-the-Diff.}
  \label{tab:gate_precision}
\end{table}

Table~\ref{tab:llmgen_precision} reports the same precision measure for the generated gates of Section~\ref{sec:experiments:llmgen}, single-shot and after the self-critique pass. $P_{\text{rand}}$ is shared with Table~\ref{tab:gate_precision}, since it depends only on the baseline error rate. These are the precision values behind the correlation in Section~\ref{sec:discussion}.

\begin{table}[H]
  \centering
  \small
  \begin{tabular}{@{}llcc@{}}
    \hline
    & \textbf{Model} & \textbf{LLM-gen} & \textbf{+Critique} \\
    \hline
    \multirow{4}{*}{CLEVR}
    & GPT-4o & 0.578 & 0.578 \\
    & Gemini & 0.549 & 0.549 \\
    & Qwen   & 0.571 & 0.571 \\
    & LLaMA  & 0.287 & 0.278 \\
    \hline
    \multirow{4}{*}{MB}
    & GPT-4o & 0.355 & 0.487 \\
    & Gemini & 0.217 & 0.353 \\
    & Qwen   & 0.521 & 0.457 \\
    & LLaMA  & 0.333 & 0.384 \\
    \hline
    \multirow{4}{*}{STD}
    & GPT-4o & 0.111 & 0.088 \\
    & Gemini & 1.000 & 0.500 \\
    & Qwen   & 0.290 & 0.289 \\
    & LLaMA  & 0.222 & 0.241 \\
    \hline
  \end{tabular}
  \caption{Gate precision for the LLM-generated gates, single-shot (LLM-gen) and self-critiqued (+Critique), all 12 cells. $P_{\text{rand}}$ as in Table~\ref{tab:gate_precision}. The Spot-the-Diff Gemini single-shot gate triggers on only 6 of 197 samples. MB = MagicBrush; STD = Spot-the-Diff.}
  \label{tab:llmgen_precision}
\end{table}

\section{Keyword Quality Analysis}
\label{app:keyword_quality}

The MagicBrush result in Section~\ref{sec:experiments:llmgen} shows that pattern quality determines whether SAVER works. Here we vary pattern quality directly. We re-implement the hand-tuned CLEVR-Change gate in a parameterized form that reproduces its per-sample decisions exactly, then degrade the evidence patterns in controlled steps and measure accuracy, trigger rate, and gate precision for GPT-4o ($N=632$). All conditions reuse cached responses.

\paragraph{Breadth.}
We broaden the patterns in four cumulative levels. L0 is the manual gate. L1 widens the proximity windows by 2.5$\times$. L2 adds bare generic change verbs (``changed'', ``different'') to the color and material categories, the same shape as the over-broad generated MagicBrush patterns. L3 adds near-universal words (``now'', ``image'', ``second'') to every category. We run each level twice: once with the conflict checks intact (CLEVR-Change's gate design) and once with them disabled, which is the structure of MagicBrush's \texttt{attribute\_change} category (Appendix~\ref{app:gate_rules}). Table~\ref{tab:keyword_breadth} shows both regimes.

\begin{table}[H]
  \centering
  \small
  \begin{tabular}{@{}llcccc@{}}
    \hline
    & \textbf{Level} & \textbf{Trig.} & \textbf{Prec.} & \textbf{Rec.} & \textbf{Acc.} \\
    \hline
    \multirow{4}{*}{\shortstack{Conflict\\checks\\intact}}
    & L0 manual  & 0.39 & 0.508 & 0.733 & 86.2 \\
    & L1 windows & 0.37 & 0.532 & 0.721 & 85.9 \\
    & L2 verbs   & 0.60 & 0.382 & 0.843 & 87.3 \\
    & L3 ubiq.   & 1.00 & 0.272 & 1.000 & 80.2 \\
    \hline
    \multirow{4}{*}{\shortstack{Conflict\\checks\\disabled}}
    & L0 manual  & 0.34 & 0.583 & 0.733 & 87.3 \\
    & L1 windows & 0.29 & 0.663 & 0.709 & 87.0 \\
    & L2 verbs   & 0.12 & 1.000 & 0.430 & 80.5 \\
    & L3 ubiq.   & 0.00 & ---   & 0.000 & 72.8 \\
    \hline
  \end{tabular}
  \caption{Breadth degradation of the hand-tuned CLEVR-Change gate (GPT-4o). Levels are cumulative. $P_{\text{rand}} = 0.272$ throughout. Reference points: B0 $=$ 72.8, B1 $=$ 80.7.}
  \label{tab:keyword_breadth}
\end{table}

\begin{figure}[H]
  \centering
  \includegraphics[width=\columnwidth]{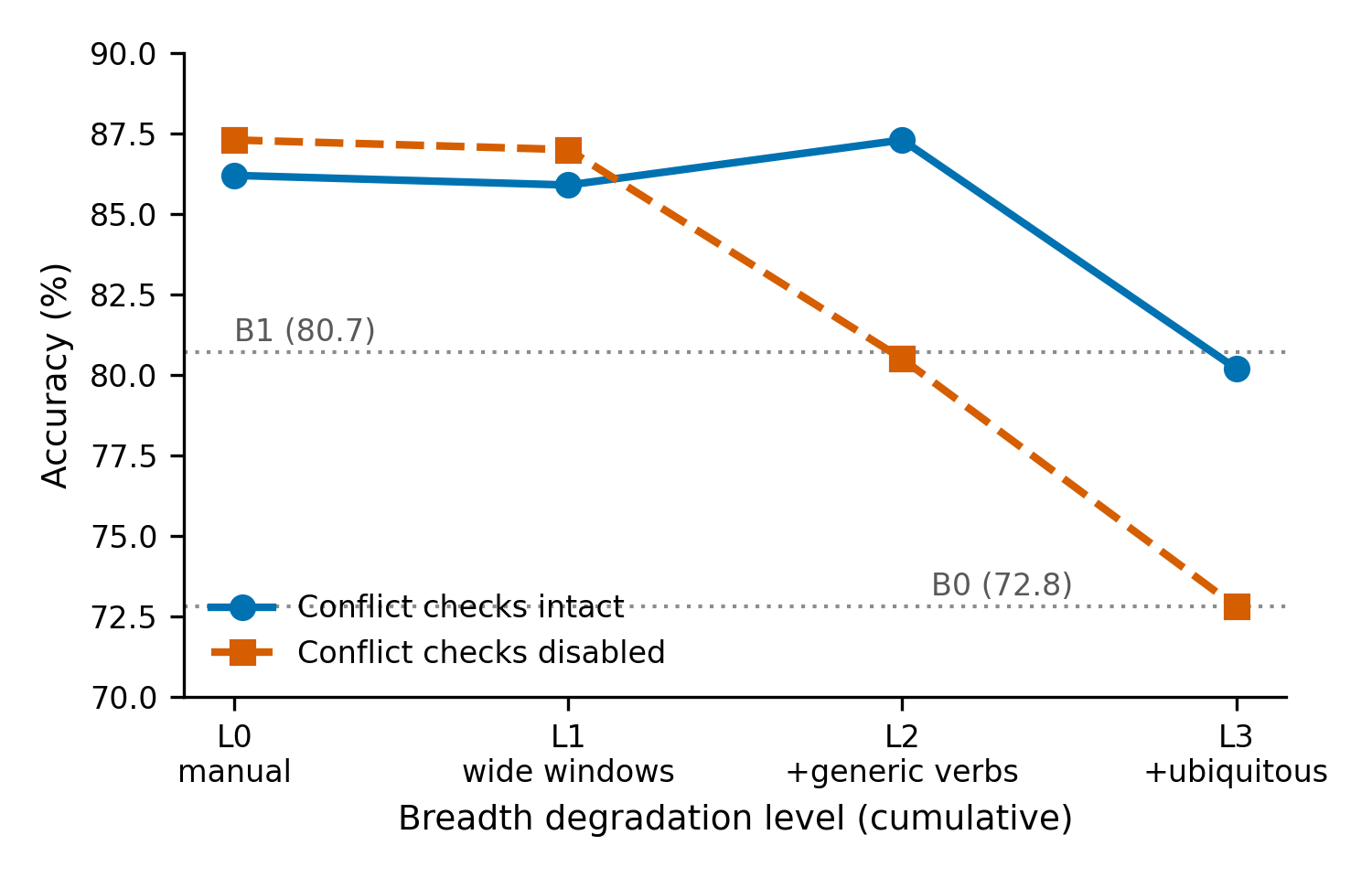}
  \caption{Breadth degradation of the hand-tuned CLEVR-Change gate (GPT-4o), both conflict-check regimes. Dotted lines mark the baseline (B0) and always-structured (B1) accuracy. With conflict checks intact, broadening over-triggers and degrades gracefully toward B1. Without them, the gate under-triggers silently and falls back to B0.}
  \label{fig:keyword_quality}
\end{figure}

The two regimes fail differently. With conflict checks intact, broad patterns make several evidence categories fire at once. The gate over-triggers, and at L3 every sample is reprompted. Accuracy degrades gracefully toward always-structured prompting (80.2 at L3, against B1 at 80.7). The cost is wasted calls, not silent failure. With conflict checks disabled, the same broadening makes the required evidence look present in nearly every response. Missing evidence becomes undetectable, the trigger rate collapses from 0.34 to zero, and accuracy falls back to the 72.8 baseline. Nothing in the gate's own output signals the failure, since the few remaining triggers have perfect precision. This is the same structure as the MagicBrush failure in Section~\ref{sec:experiments:llmgen}, reproduced in a controlled setting. Figure~\ref{fig:keyword_quality} plots both regimes.

\paragraph{Coverage.}
We remove 25\%, 50\%, or 75\% of the pattern branches in each evidence category (three random draws, seed 42, at least one branch kept per category). Table~\ref{tab:keyword_coverage} shows the result. Missing keywords push the gate the other way. Evidence looks absent more often, the gate over-triggers, precision falls toward $P_{\text{rand}}$, and accuracy stays in the 81.6--86.6 band, well above baseline. Missing keywords are therefore the graceful failure mode. Over-broad keywords without conflict checks are the silent one.

\begin{table}[H]
  \centering
  \small
  \begin{tabular}{@{}lccc@{}}
    \hline
    \textbf{Dropped} & \textbf{Trig.} & \textbf{Prec.} & \textbf{Acc.} \\
    \hline
    0\% (manual) & 0.39 & 0.508 & 86.2 \\
    25\% & 0.66 $\pm$ 0.09 & 0.36 $\pm$ 0.04 & 82.8 $\pm$ 1.7 \\
    50\% & 0.53 $\pm$ 0.06 & 0.44 $\pm$ 0.05 & 86.6 $\pm$ 3.0 \\
    75\% & 0.92 $\pm$ 0.10 & 0.29 $\pm$ 0.02 & 81.6 $\pm$ 1.7 \\
    \hline
  \end{tabular}
  \caption{Coverage degradation of the hand-tuned CLEVR-Change gate (GPT-4o). Mean $\pm$ sd over three draws. $P_{\text{rand}} = 0.272$.}
  \label{tab:keyword_coverage}
\end{table}

\section{Cost Comparison}
\label{app:cost}

Table~\ref{tab:cost} compares SAVER to always-structured (B1) and self-consistency (SC-5, GPT-4o only). On CLEVR-Change, SAVER outperforms B1 for GPT-4o (+5.5\%) and LLaMA (+2.7\%) while using only 1.37--1.39 calls, because the gate avoids structured prompting where it hurts. SC-5 achieves 77.4\% at 5.0 calls/sample, below both SAVER (86.2\% at 1.39 calls) and B1 (80.7\% at 1.0 calls), making it the least cost-effective strategy on this dataset. For Gemini and Qwen, B1 is more accurate, meaning always structured is the better strategy. On MagicBrush, B1 outperforms SAVER for all models by 7--13\%, because errors there are predominantly perceptual and benefit from universal structured prompting. Thus, SAVER works best when structured prompting helps some input types but hurts others.

\begin{table}[H]
  \centering
  \small
  \begin{tabular}{@{}llcccc@{}}
    \hline
    & \textbf{Model} & \textbf{B1} & \textbf{SC-5} & \textbf{SAVER} & \textbf{Calls} \\
    \hline
    \multirow{4}{*}{CLEVR}
    & GPT-4o & 80.7 & 77.4 & \textbf{86.2} & 1.39 \\
    & Gemini & \textbf{95.6} & --- & 93.0 & 1.36 \\
    & Qwen   & \textbf{87.0} & --- & 82.3 & 1.49 \\
    & LLaMA  & 86.2 & --- & \textbf{88.9} & 1.37 \\
    \hline
    \multirow{4}{*}{MB}
    & GPT-4o & \textbf{70.3} & --- & 62.9 & 1.22 \\
    & Gemini & \textbf{79.7} & --- & 68.8 & 1.11 \\
    & Qwen   & \textbf{68.8} & --- & 56.2 & 1.38 \\
    & LLaMA  & \textbf{74.4} & --- & 63.5 & 1.30 \\
    \hline
  \end{tabular}
  \caption{Accuracy (\%) vs.\ API cost. B1 uses 1.0 calls; SAVER averages 1.11--1.49 (Calls column); SC-5 uses 5.0 calls (GPT-4o CLEVR only). \textbf{Bold} = highest accuracy.}
  \label{tab:cost}
\end{table}

\section{Ablation Details}
\label{app:ablation}

Beyond the GPT-4o ablations in the main paper, we tested three additional gate configurations (Table~\ref{tab:gate_sensitivity}).

\paragraph{C5: Non-critical keyword trigger.}
C5 replaces SAVER's evidence-based trigger with spatial and background keywords (e.g., \emph{left}, \emph{right}, \emph{behind}, \emph{center}). The trigger count is matched to SAVER's so the two conditions differ only in \emph{which} samples are reprompted. On CLEVR-Change, C5 underperforms SAVER by 4--8\% across all four models (Table~\ref{tab:gate_sensitivity}). The gap is largest for GPT-4o (78.5\% vs.\ 86.2\%). Spatial keywords appear in both correct and incorrect responses, so C5 reprompts many samples that do not need it while missing samples that do. On MagicBrush the gap narrows, and LLaMA is the one exception where C5 slightly outperforms SAVER (64.1\% vs.\ 63.5\%). This result shows that evidence-type matching, not just keyword presence, is what makes the gate effective.

\paragraph{C6: Strict gate.}
C6 triggers whenever \emph{any} evidence dimension is missing, regardless of the predicted change type. This is the most aggressive variant: trigger rates reach 88--100\% on CLEVR and 55--79\% on MagicBrush, meaning nearly every sample gets reprompted. On CLEVR, C6 approximates always-structured (B1) and scores below SAVER for GPT-4o and LLaMA, the two models where B1 hurts drop/add types. For Gemini and Qwen, C6 slightly exceeds SAVER because these models benefit from structured reasoning on all input types. On MagicBrush, C6 outperforms SAVER for all four models. MagicBrush errors are predominantly perceptual, so broader triggering compensates: even when the gate cannot distinguish expression from perception failures, the structured prompt still helps by forcing more careful descriptions of real photographs.

\paragraph{C7: Loose gate.}
C7 triggers only when conflicting evidence is detected (e.g., the response claims ``color change'' but contains material-related terms). This is the most conservative variant, with trigger rates of 5--33\%. On CLEVR, C7 scores below SAVER by 4--16\% across all models. The largest drop is for Qwen (66.0\% vs.\ 82.3\%), where many incorrect baselines simply lack evidence rather than containing conflicting evidence, so C7 misses them. On MagicBrush, C7 performs close to baseline for most models. The contrast between C6 and C7 shows that SAVER's standard gate occupies a productive middle ground: strict enough to catch evidence-poor responses, but not so strict that it reprompts samples the model already handles well.

\begin{table*}[!ht]
  \centering
  \small
  \begin{tabular}{@{}lccccccccc@{}}
    \hline
    & & \multicolumn{4}{c}{\textbf{CLEVR-Change}} & \multicolumn{4}{c}{\textbf{MagicBrush}} \\
    \textbf{Condition} & \textbf{Trig.} & \textbf{GPT-4o} & \textbf{Gemini} & \textbf{Qwen} & \textbf{LLaMA} & \textbf{GPT-4o} & \textbf{Gemini} & \textbf{Qwen} & \textbf{LLaMA} \\
    \hline
    SAVER (ref)     & 11--49\%  & \textbf{86.2} & 93.0          & 82.3          & \textbf{88.9} & 62.9          & 68.8          & 56.2          & 63.5 \\
    C5 Non-critical & matched   & 78.5          & 86.4          & 76.3          & 84.5          & 61.5          & 69.1          & 52.4          & 64.1 \\
    C6 Strict gate  & 55--100\% & 81.8          & \textbf{95.6} & \textbf{87.2} & 87.0          & \textbf{66.8} & \textbf{77.1} & \textbf{64.7} & \textbf{67.4} \\
    C7 Loose gate   & 5--33\%   & 81.5          & 88.0          & 66.0          & 85.0          & 61.5          & 69.1          & 55.0          & 59.4 \\
    \hline
  \end{tabular}
  \caption{Gate design sensitivity analysis (\% accuracy). C5 = non-critical keyword trigger; C6 = strict gate (any missing evidence); C7 = loose gate (conflict only). \textbf{Bold} = best per column. Spot-the-Diff is excluded (no B1 structured data for these conditions).}
  \label{tab:gate_sensitivity}
\end{table*}

\paragraph{Self-consistency comparison.}
SC-5 scores below both SAVER and B1 on CLEVR-Change despite 3.6$\times$ more API calls (Table~\ref{tab:ablation}). The per-type breakdown (Table~\ref{tab:clevr_pertype}) explains why: SC-5 barely helps texture, where GPT-4o consistently mislabels texture as color. Majority voting amplifies this confusion. SAVER sidesteps the problem by checking for material/texture keywords directly, correcting 88 samples that SC-5 misses while SC-5 corrects only 32 that SAVER misses.

SAVER averages $1 + \text{trigger\_rate}$ calls per sample. For GPT-4o on CLEVR-Change (39\% trigger rate), this is 1.39 calls, a 39\% overhead that yields +13.4\% accuracy. Yet SC-5 requires 5 calls for a smaller gain. The full cost--accuracy comparison is in Appendix~\ref{app:cost}.

\section{Per-Type Visualization}
\label{app:pertype_fig}

\begin{figure}[H]
  \centering
  \includegraphics[width=\columnwidth]{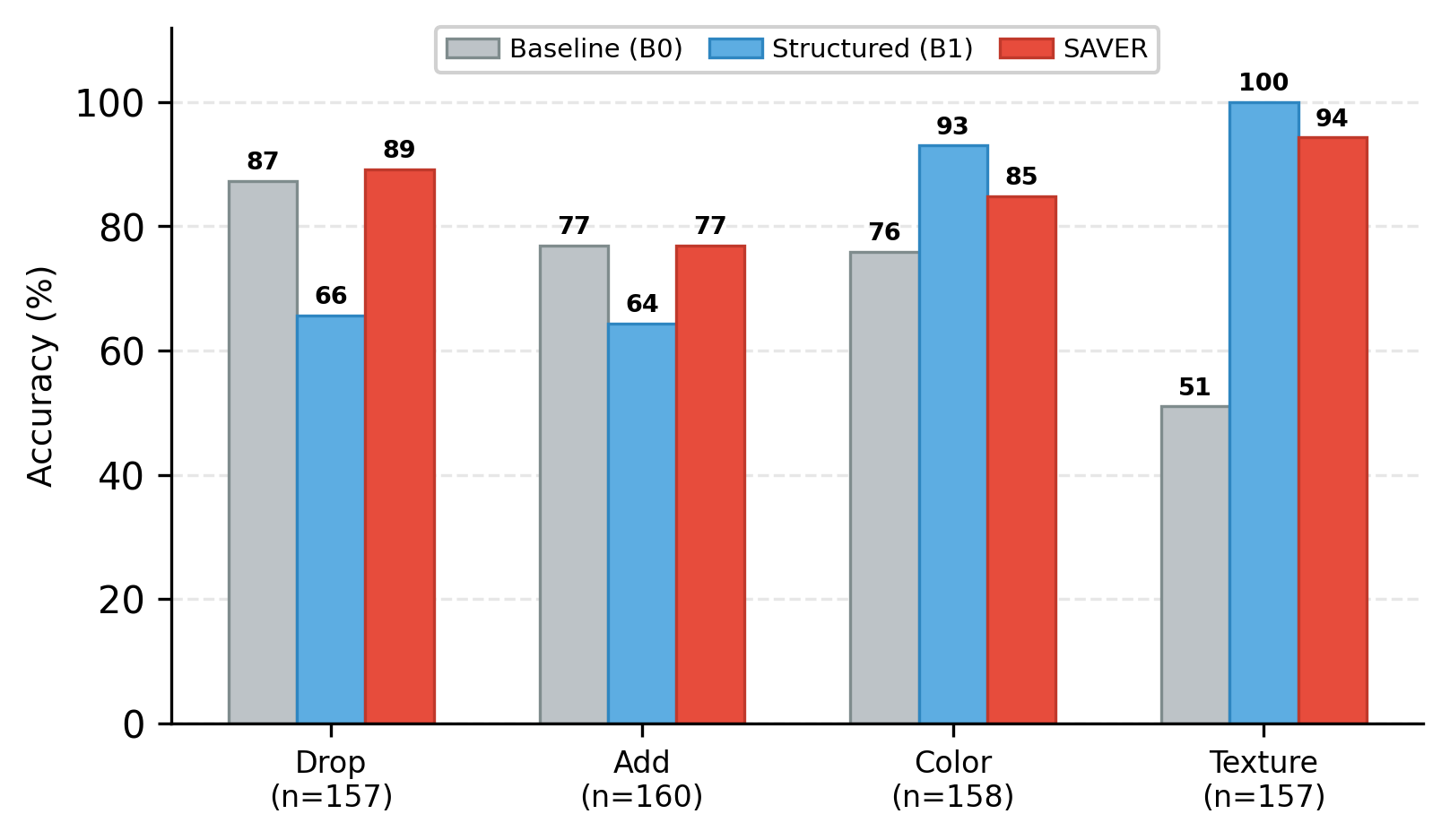}
  \caption{CLEVR-Change per-type accuracy for GPT-4o. Structured prompting (B1) hurts drop and add types but helps color and texture. SAVER selectively applies structured reasoning where it helps.}
  \label{fig:pertype}
\end{figure}

Figure~\ref{fig:pertype} visualizes the per-type accuracy breakdown from Table~\ref{tab:clevr_pertype} in the main paper.

\section{Exploratory: Claude 3.5 Sonnet}
\label{app:claude}

We ran Claude 3.5 Sonnet \cite{anthropic2024claude35} on a 200-sample CLEVR-Change subset as an exploratory comparison. Table~\ref{tab:claude} shows the results. This is one of the cleanest cases for selective gating: SAVER outperforms B1 on accuracy (77.5\% vs.\ 76.5\%) while breaking zero correct baselines. B1 fixes 27 samples but breaks 12 (6.0\% broke rate), yielding a net gain of +15. SAVER fixes 17 and breaks 0, for a net gain of +17. SAVER achieves both higher accuracy and better robustness simultaneously because the gate avoids all collateral damage from unnecessary reprompting.

\begin{table}[H]
  \centering
  \small
  \begin{tabular}{@{}lcccc@{}}
    \hline
    \textbf{Condition} & \textbf{Accuracy} & \textbf{Fix} & \textbf{Broke} & \textbf{Trigger} \\
    \hline
    Baseline (B0)   & 69.0\% & --- & --- & --- \\
    Structured (B1) & 76.5\% & 27  & 12  & --- \\
    SAVER           & \textbf{77.5\%} & 17  & 0   & 20\% \\
    \hline
  \end{tabular}
  \caption{Claude 3.5 Sonnet on CLEVR-Change (200 samples). \textbf{Bold} = best. SAVER outperforms B1 with zero breaks.}
  \label{tab:claude}
\end{table}

\section{Domain Applicability}
\label{app:domains}

SAVER applies when a task has a structured output taxonomy and known evidence requirements per output type. Change detection satisfies both: change types (drop, add, color, texture) form the taxonomy, and each type requires specific verbal evidence (absence language, color transition phrases). We describe three other domains where these conditions hold and what adaptation would involve.

\paragraph{Manufacturing defect inspection.}
Defect categories such as scratch, dent, discoloration, and missing component form a taxonomy. Each category maps to visual evidence: scratches require directional terms (``along the edge,'' ``linear mark''), discoloration requires color deviation terms (``darker region,'' ``staining''). A gate would check whether the VLM's defect report contains the evidence that the claimed category requires. In manufacturing, a missed defect is typically far costlier than a false alarm. This favors higher trigger rates, closer to our C6 strict-gate variant (Appendix~\ref{app:ablation}), where the gate reprompts aggressively and accepts the cost of redundant calls.

\paragraph{Medical image comparison.}
Comparing longitudinal medical images (e.g., chest X-rays or MRI scans over time) involves detecting changes in lesion size, new findings, or resolved conditions. The output taxonomy includes categories like new lesion, growth, shrinkage, and stable. A ``growth'' claim should reference size measurements or comparative terms (``larger,'' ``increased in diameter''), giving clear evidence requirements. Two challenges separate this domain from change detection. First, subtle changes may fall below VLM perceptual thresholds, so errors would be perceptual rather than expressive. Second, image registration artifacts can produce spurious visual differences. Our Spot-the-Diff results, where perception failures dominate, suggest SAVER would provide less benefit in such settings unless paired with domain-specific preprocessing to reduce perceptual noise.

\paragraph{Document change tracking.}
Comparing document versions (contracts, regulatory filings, technical specifications) is a structured comparison task with clear change types: additions, deletions, modifications, and reformulations. A ``modification'' claim should cite the specific text that changed and how. Because document comparison is predominantly textual, the perception failure mode that limits SAVER on natural images is less relevant. This makes document comparison a strong candidate for evidence gating.

\paragraph{General adaptation.}
Applying SAVER to a new domain requires four steps: (1) define the output taxonomy, (2) specify required evidence keywords per type, (3) optionally define conflicting evidence pairs, and (4) validate the gate against the random-gate baseline on a small labeled sample (Appendix~\ref{app:gate_precision}). Steps 1--3 can be automated by prompting an LLM with a task description and example outputs (Section~\ref{sec:experiments:llmgen}). The main requirement is that the domain exhibits expression failures, where the model has the right information but omits it from its output. Domains where errors are mostly perceptual will benefit less, as our results on Spot-the-Diff confirm, and on MagicBrush to a lesser degree.

\section{Spot-the-Diff Case Study: Perception Failures}
\label{app:spotdiff_cases}

To verify that Spot-the-Diff errors are perceptual rather than expressive, we examined gate-accepted responses that were scored as incorrect. Table~\ref{tab:perception_rate} shows the perception failure rate across all four models: 27--46\% of gate-accepted responses are wrong despite containing evidence language. The gate finds change-related keywords (disappearance, movement, appearance) and accepts the response, but the described changes do not match the ground truth. Note that these accepted-but-wrong cases do not appear in the Fix/Broke counts (Table~\ref{tab:fixbroke}), which only count samples where SAVER changed the outcome through reprompting. The cases here were never reprompted.

\begin{table}[H]
  \centering
  \small
  \begin{tabular}{@{}lccc@{}}
    \hline
    \textbf{Model} & \textbf{Accepted} & \textbf{Wrong} & \textbf{Perc.\ Fail.\ \%} \\
    \hline
    GPT-4o  & 169 & 45 & 26.6 \\
    Gemini  & 163 & 61 & 37.4 \\
    Qwen    & 173 & 79 & 45.7 \\
    LLaMA   &  84 & 28 & 33.3 \\
    \hline
  \end{tabular}
  \caption{Perception failure rates on Spot-the-Diff. ``Accepted'' = gate found evidence and did not trigger. ``Wrong'' = accepted but incorrect. These are cases where the model describes changes confidently but inaccurately.}
  \label{tab:perception_rate}
\end{table}

Three GPT-4o examples illustrate the pattern. In each case, the gate accepts the response because evidence keywords are present, but the described changes do not match the ground truth.

\begin{tcolorbox}[colback=yellow!8, colframe=yellow!50!black, title={\small Example 1 (ID 10313)}, fonttitle=\bfseries\small, boxrule=0.4pt, left=4pt, right=4pt, top=2pt, bottom=2pt]
\small
\textbf{Ground truth:} the red car is no longer there. \\
\textbf{GPT-4o:} A white car moved forward, a black car shifted right, a person appeared. \\
\textbf{Gate:} Accepts (movement + appearance evidence found). None of the three described changes match the actual disappearance.
\end{tcolorbox}

\begin{tcolorbox}[colback=yellow!8, colframe=yellow!50!black, title={\small Example 2 (ID 8648)}, fonttitle=\bfseries\small, boxrule=0.4pt, left=4pt, right=4pt, top=2pt, bottom=2pt]
\small
\textbf{Ground truth:} Several people are missing; a few new people appear. \\
\textbf{GPT-4o:} A cyclist disappeared; a walking person shifted left. \\
\textbf{Gate:} Accepts (disappearance + movement evidence found). The response describes the wrong people and wrong changes.
\end{tcolorbox}

\begin{tcolorbox}[colback=yellow!8, colframe=yellow!50!black, title={\small Example 3 (ID 9900)}, fonttitle=\bfseries\small, boxrule=0.4pt, left=4pt, right=4pt, top=2pt, bottom=2pt]
\small
\textbf{Ground truth:} A person appeared by the dumpster; two background people disappeared; a man in yellow appeared. \\
\textbf{GPT-4o:} A white car disappeared; a silver car moved forward. \\
\textbf{Gate:} Accepts (disappearance + movement evidence found). The actual changes involve people, but the model describes vehicles.
\end{tcolorbox}

\noindent The model produces confident, evidence-rich descriptions in all three cases. The gate cannot catch these errors because the responses contain valid evidence keywords. The failures are perceptual, not expressive.

\section{Hyperparameters and Reproducibility}
\label{app:reproducibility}

All experiments use the following settings:

\begin{itemize}
    \item \textbf{Temperature}: 0 for all conditions except self-consistency ($k$=5, temperature 0.7)
    \item \textbf{Max tokens}: 300 (baseline), 600 (structured/SAVER reprompt), 400 (standard CoT)
    \item \textbf{Image detail}: \texttt{low} for GPT-4o, \texttt{auto} for OpenRouter models. Images are base64-encoded PNG.
    \item \textbf{Retry policy}: up to 3 retries with exponential backoff (30--120\,s) on rate-limit or server errors
    \item \textbf{Random seed}: 42 for all stochastic operations (sample selection, bootstrap, ablation randomization)
    \item \textbf{Bootstrap}: 1,000 resamples, 95\% confidence intervals
    \item \textbf{Significance testing}: McNemar's test with continuity correction, Bonferroni adjusted across 12 model--dataset pairs. The LLM-generated pattern and self-critique experiments form a separate test family with their own Bonferroni correction (24 tests). The original tests and their significance markers are unchanged.
\end{itemize}

\noindent Table~\ref{tab:model_versions} lists the exact model IDs and API providers. The main experiments ran between January and February 2026. The LLM-generated pattern and self-critique experiments ran in July 2026. They reuse the cached January--February responses wherever a gate decision matches an already-run condition. Only 38 new Spot-the-Diff structured reprompts were needed: 28 for GPT-4o and 10 for LLaMA-4-Scout. By July 2026, the original GPT-4o endpoint was replaced by OpenRouter's \texttt{openai/gpt-4o} (same underlying model, image detail \texttt{low} as before). Gemini 2.0 Flash and Qwen2.5-VL-7B had been retired by their providers before the revision experiments, so their Spot-the-Diff cells rely on cached responses only. Pattern generation and self-critique used Claude Sonnet 4.5, which is not among the evaluated VLMs. Per dataset, single-shot pattern generation cost \$0.02--\$0.04 and 16--28 seconds, and the self-critique pass cost \$0.02--\$0.05 and 14--29 seconds.

\begin{table}[H]
  \centering
  \scriptsize
  \begin{tabular}{@{}ll@{}}
    \hline
    \textbf{Model} & \textbf{API / Model ID} \\
    \hline
    GPT-4o           & OpenAI: \texttt{gpt-4o} \\
    Gemini 2.0 Flash & OpenRouter: \texttt{gemini-2.0-flash-001}$^\dagger$ \\
    Qwen2.5-VL-7B   & OpenRouter: \texttt{qwen-2.5-vl-7b-instruct}$^\dagger$ \\
    LLaMA-4-Scout    & OpenRouter: \texttt{llama-4-scout} \\
    Claude 3.5 Sonnet & OpenRouter: \texttt{claude-3.5-sonnet} \\
    \hline
    GPT-4o (revision)     & OpenRouter: \texttt{openai/gpt-4o} \\
    Claude Sonnet 4.5 & Anthropic: \texttt{claude-sonnet-4-5} \\
    \hline
  \end{tabular}
  \caption{Model versions and API endpoints. Top block: main experiments, January--February 2026. Bottom block: revision experiments, July 2026 (28 new GPT-4o Spot-the-Diff reprompts via OpenRouter; Claude Sonnet 4.5 as pattern generator and critic). $^\dagger$ = retired by the provider before the revision experiments.}
  \label{tab:model_versions}
\end{table}

\end{document}